\documentclass{article}

\usepackage{microtype}
\usepackage{graphicx}
\usepackage{subfigure}
\usepackage{booktabs}
\usepackage{hyperref}

\usepackage[accepted]{icml2019}

\icmltitlerunning{Generating Long Sequences with Sparse Transformers}

\begin{document}

\twocolumn[
\icmltitle{Generating Long Sequences with Sparse Transformers}

\icmlsetsymbol{equal}{*}

\begin{icmlauthorlist}
\icmlauthor{Rewon Child}{openai}
\icmlauthor{Scott Gray}{openai}
\icmlauthor{Alec Radford}{openai}
\icmlauthor{Ilya Sutskever}{openai}
\end{icmlauthorlist}

\icmlaffiliation{openai}{OpenAI, San Francisco, California, United States}

\icmlcorrespondingauthor{Rewon Child}{rewon@openai.com}

\icmlkeywords{Machine Learning, self-attention, attention, generative modeling}

\vskip 0.3in
]

\begin{abstract}
Transformers are powerful sequence models, but require time and memory that grows quadratically with the sequence length. In this paper we introduce sparse factorizations of the attention matrix which reduce this to $O(n \sqrt{n})$. We also introduce a) a variation on architecture and initialization to train deeper networks, b) the recomputation of attention matrices to save memory, and c) fast attention kernels for training. We call networks with these changes Sparse Transformers, and show they can model sequences tens of thousands of timesteps long using hundreds of layers. We use the same architecture to model images, audio, and text from raw bytes, setting a new state of the art for density modeling of Enwik8, CIFAR-10, and ImageNet-64. We generate unconditional samples that demonstrate global coherence and great diversity, and show it is possible in principle to use self-attention to model sequences of length one million or more. \end{abstract}

\section{Introduction}
\label{introduction}

Estimating complex, high-dimensional data distributions is a central problem in unsupervised learning, as many downstream applications of interest involve generation of text, images, audio, and other data. Additionally, it is believed to be a key component of unsupervised representation learning.

Recently, neural autoregressive models have achieved impressive results in this domain, achieving state-of-the-art in modeling natural language \cite{jozefowicz2016exploring} \cite{radford2018} \cite{dai2018transformer}, raw audio \cite{van2016wavenet} \cite{mehri2016samplernn}, and images \cite{oord2016pixel} \cite{menick2018generating} \cite{salimans2017pixelcnn++} \cite{reed2017parallel} \cite{chen2017pixelsnail}.

These methods decompose a joint probability distribution into a product of conditional ones. Modeling these conditional distributions is extremely challenging, however, as they contain many complex, long-range dependencies and require a suitably expressive model architecture to learn them.

\begin{figure}[t]
\centering
\setlength\fboxsep{0pt}
\setlength\fboxrule{0.25pt}
\includegraphics[width=0.46\textwidth]{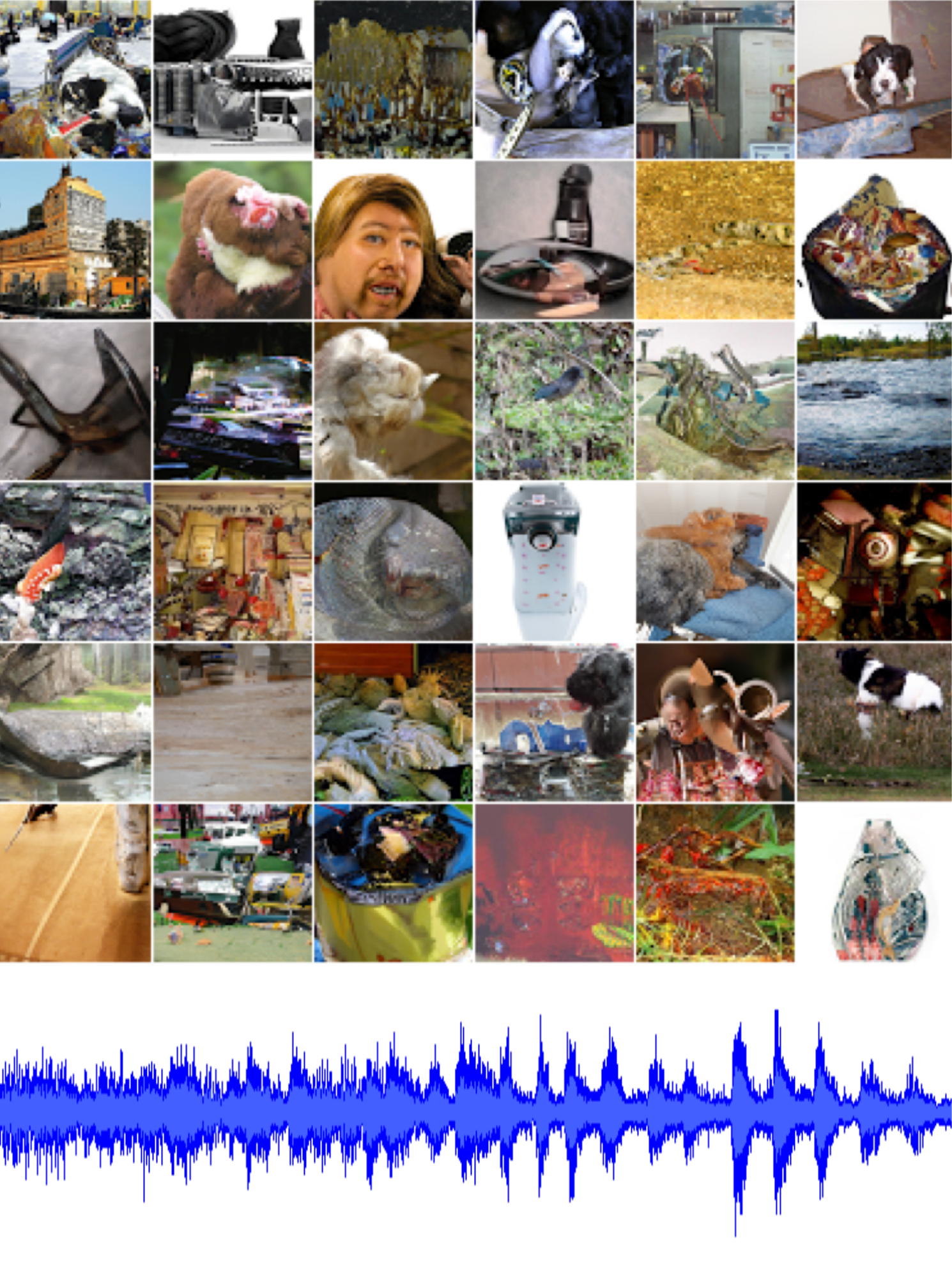}
\caption{Unconditional samples from our neural autoregressive model on ImageNet 64 and a classical music dataset. We used the same self-attention based architecture for audio, images, and text. The samples above were generated with softmax temperature 1.0, and had lengths 12,288 and 65,536. Audio samples be listened to at \url{https://openai.com/blog/sparse-transformer}}
\label{uncond_samples}
\end{figure}

Architectures based off CNNs \cite{oord2016pixel} have made great progress in this direction, but require significant depth to expand their receptive field. To address this, WaveNet \cite{van2016wavenet} introduced dilated convolutions, which allowed the network to model long-range dependencies in a logarithmic number of layers.

Separately, the Transformer \cite{vaswani2017attention} has been shown to excel on many natural language tasks, which may be in part due to its ability to model arbitrary dependencies in a constant number of layers. As each self-attention layer has a global receptive field, the network can allocate representational capacity to the input regions for which it is most useful. Thus the architecture may be more flexible at generating diverse data types than networks with fixed connectivity patterns.

However, the memory and computational requirements of such networks grows quadratically with sequence length, which excludes their use on long sequences.

\begin{figure*}
\centering     
\includegraphics[width=0.95\textwidth]{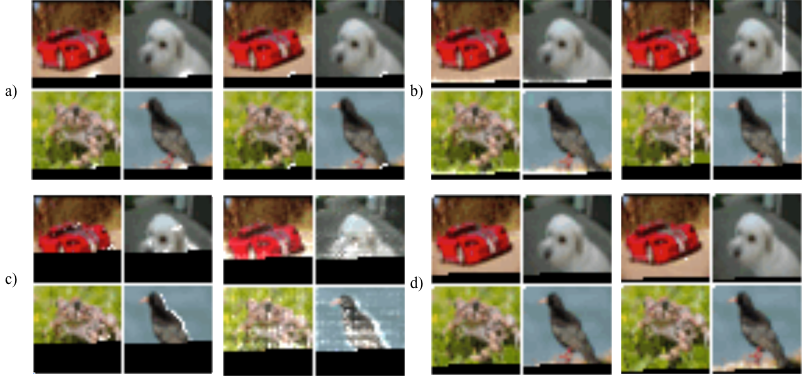}
\caption{Learned attention patterns from a 128-layer network on CIFAR-10 trained with full attention. White highlights denote attention weights for a head while generating a given pixel, and black denotes the autoregressive mask. Layers are able to learn a variety of specialized sparse structures, which may explain their ability to adapt to different domains. a) Many early layers in the network learn locally connected patterns, which resemble convolution. b) In layers 19 and 20, the network learned to split the attention across a row attention and column attention, effectively factorizing the global attention calculation. c) Several attention layers showed global, data-dependent access patterns. d) Typical layers in layers 64-128 exhibited high sparsity, with positions activating rarely and only for specific input patterns.}
\label{learned_patterns}
\end{figure*}

The main contribution of this work is to introduce several sparse factorizations of the attention matrix, which scale as $O(n \sqrt[p]{n})$ with the sequence length without sacrificing performance. These work by separating the full attention computation into several faster attention operations which, when combined, can approximate the dense attention operation. We use this to apply self-attention to sequences of unprecedented length.

Additionally, we introduce several other changes to the Transformer, including:

\begin{itemize}
  \item A restructured residual block and weight initialization to improve training of very deep networks
  \item A set of sparse attention kernels which efficiently compute subsets of the attention matrix
  \item Recomputation of attention weights during the backwards pass to reduce memory usage
\end{itemize}

We empirically validate that models augmented in this manner can achieve state-of-the-art compression and generation of natural language, raw audio, and natural images. The simplicity of the architecture leads us to believe it may be useful for many problems of interest.

\begin{figure*}
\centering     
\subfigure[\small{Transformer}]{\label{patt:a}\includegraphics[width=45mm]{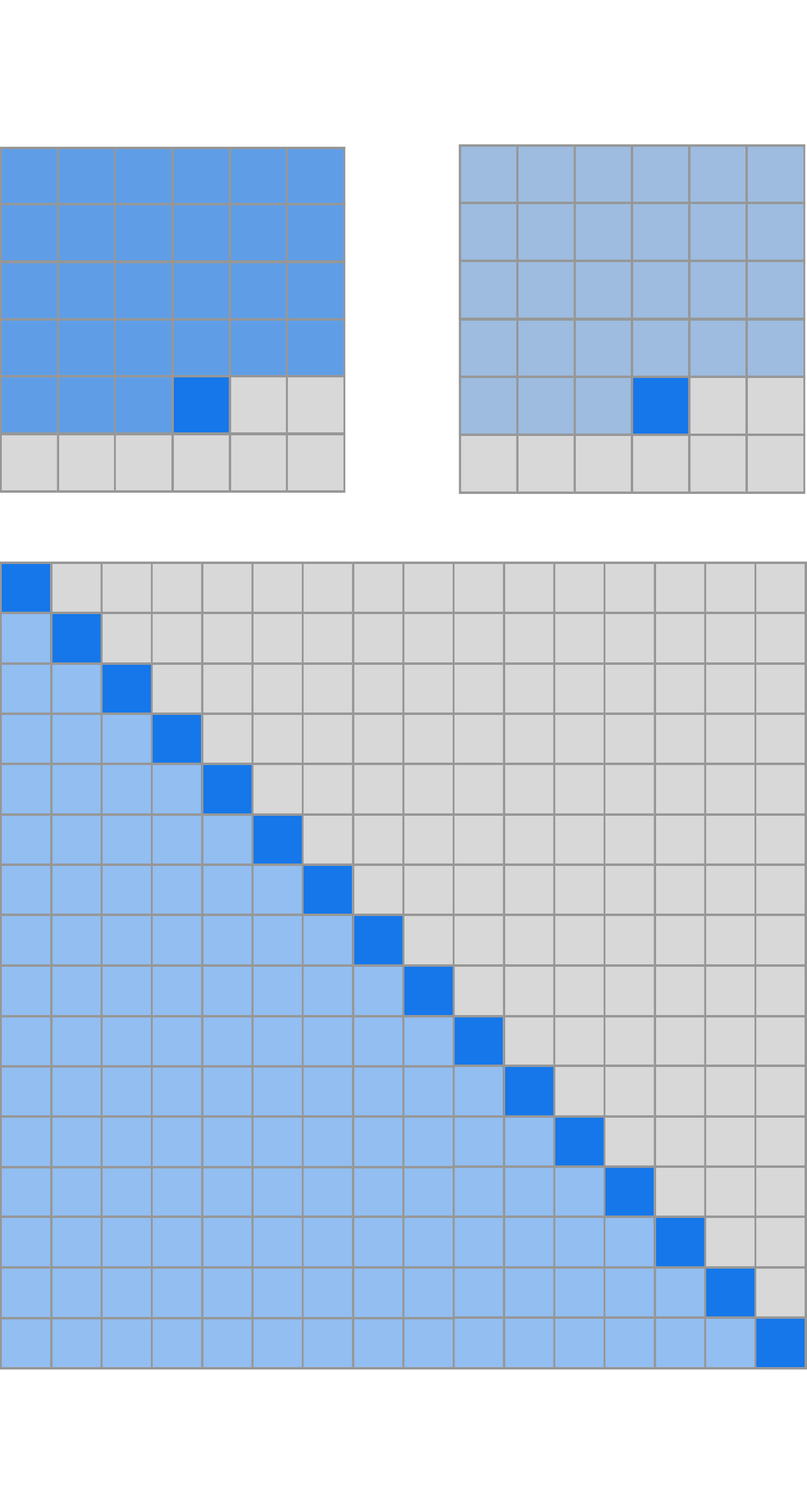}} \qquad \qquad
\subfigure[\small{Sparse Transformer (strided)}]{\label{patt:b}\includegraphics[width=45mm]{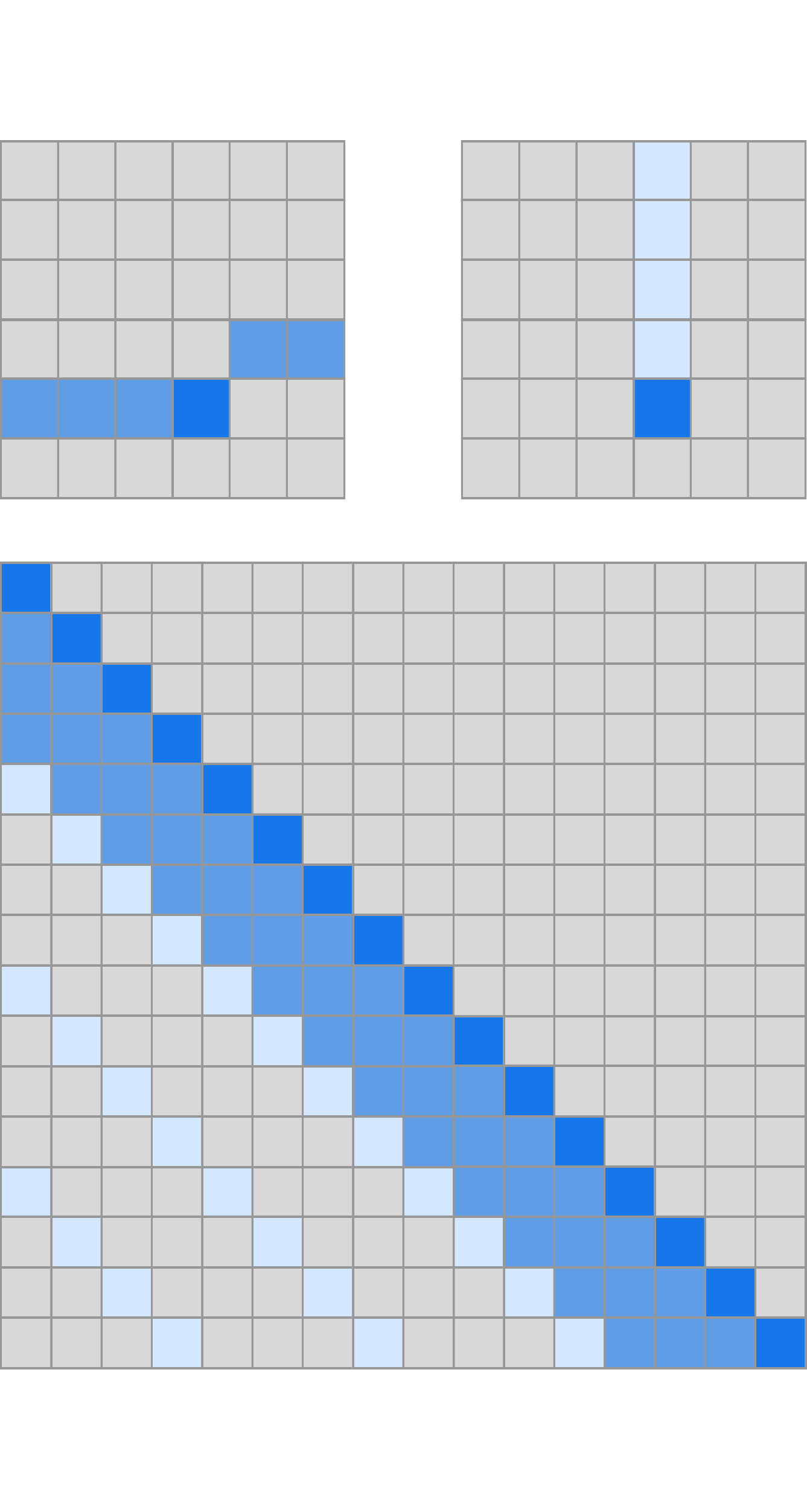}} \qquad \qquad
\subfigure[\small{Sparse Transformer (fixed)}]{\label{patt:c}\includegraphics[width=45mm]{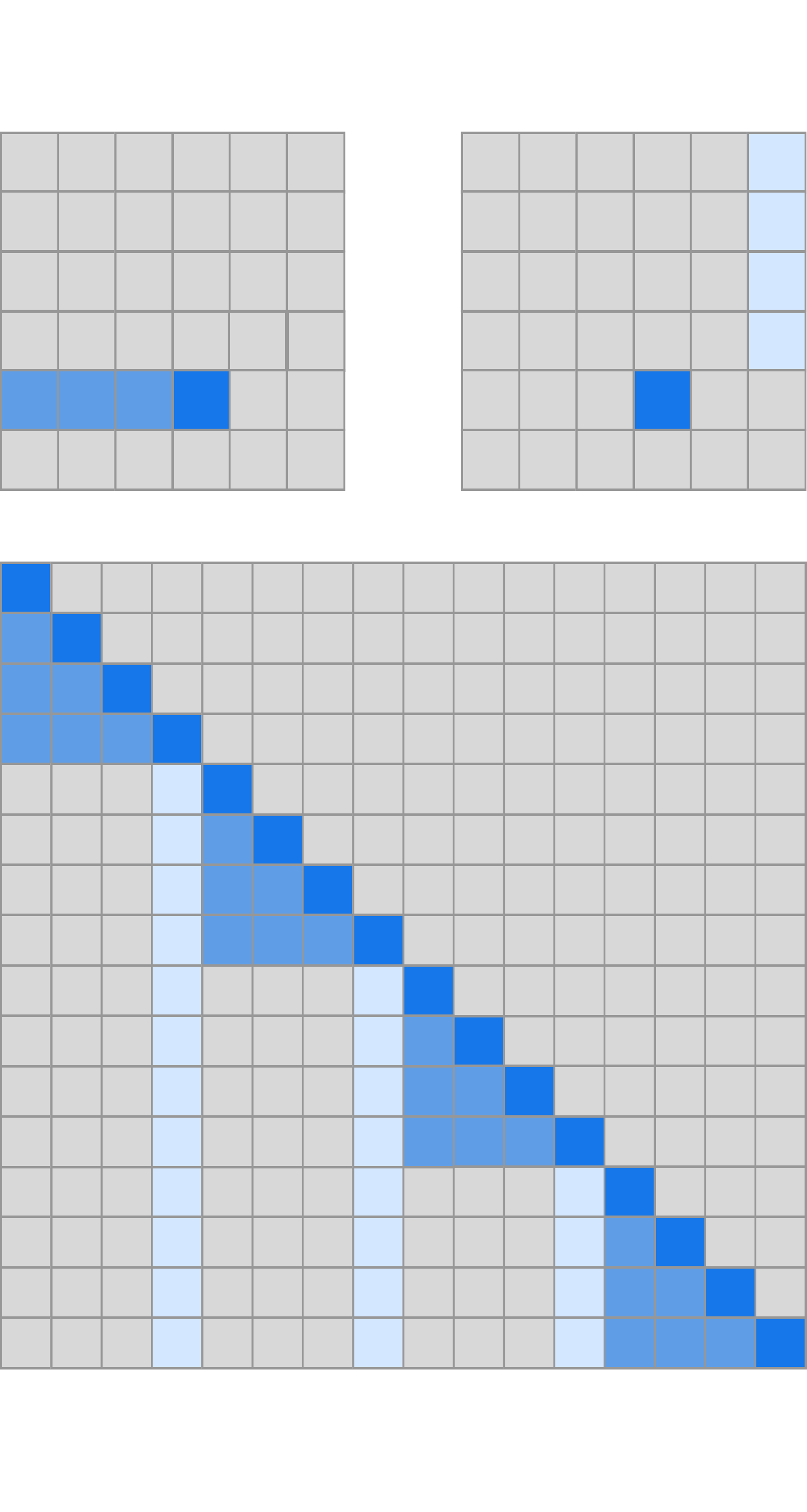}}
\caption{Two 2d factorized attention schemes we evaluated in comparison to the full attention of a standard Transformer (a). The top row indicates, for an example 6x6 image, which positions two attention heads receive as input when computing a given output. The bottom row shows the connectivity matrix (not to scale) between all such outputs (rows) and inputs (columns). Sparsity in the connectivity matrix can lead to significantly faster computation. In (b) and (c), full connectivity between elements is preserved when the two heads are computed sequentially. We tested whether such factorizations could match in performance the rich connectivity patterns of Figure \ref{learned_patterns}. }
\end{figure*}

\section{Related Work}

The most related work involves other techniques for scaling up autoregressive generative models. For images, \cite{reed2017parallel} models conditional independence between the pixels in order to generate many locations in parallel, and \cite{menick2018generating} imposes an ordering and multi-scale upsampling procedure to generate high fidelity samples. \cite{parmar2018image} uses blocks of local attention to apply Transformers to images.  For text, \cite{dai2018transformer} introduces a state reuse "memory" for modeling long-term dependencies. And for audio, in addition to \cite{van2016wavenet}, \cite{mehri2016samplernn} used a hierarchical structure and RNNs of varying clock-rates to use long contexts during inference, similar to \cite{koutnik2014clockwork}. \cite{huang2018improved} apply Transformers to MIDI generation with an efficient relative attention.

Our work is simpler than many of the techniques above and can be applied equally across images, text, and audio. Many of the above techniques are orthogonal to ours, moreover, and could be used in conjunction with ours.

Outside of generative modeling, there are several works relevant to improving the efficiency of attention based off chunking \cite{chiu2017monotonic} or using fixed length representations \cite{britz2017efficient}. Other works have investigated attention with multiple "hops", such as \cite{sukhbaatar2015end} and  \cite{gehring2017convolutional}.

It is worth noting that the Gated Pixel CNN \cite{oord2016pixel} and WaveNet \cite{van2016wavenet} use multiplicative interactions in their networks, which are related to self-attention.

\section{Background}
We consider the task of autoregressive sequence generation, where the joint probability of a sequence $\mathbf{x} = \{x_1, x_2, ..., x_n\}$ is modeled as the product of conditional probability distributions and parameterized by a network $\theta$.

\begin{equation}\label{autoregressive}
p(\mathbf{x}) = \prod_{i=1}^{n} p(x_i \vert x_1, ..., x_{i-1}; \theta) \end{equation}

We treat images, text, and audio as a sequence of discrete tokens, typically raw bytes. The network $\theta$ takes in the sequence of tokens and outputs a categorical distribution over the $v$ possible values of the next token using the $\mathrm{softmax}$ function, where $v$ is the size of the \textit{vocabulary}. The training objective is to maximize the log-probability of the data with respect to $\theta$.

A simple and powerful choice for model $\theta$ is a Transformer \cite{vaswani2017attention} in decoder-only mode, as demonstrated by \cite{radford2018} and \cite{liu2018generating}. These models transform the input sequence with blocks of multihead self-attention over the entire sequence, followed by dense transformations over each sequence element. The self-attention portion of the network must compute $n$ weightings for each of $n$ elements, however, which can quickly become intractable as the sequence length grows.

In the following sections, we describe our modifications to the Transformer architecture which make it more suitable for modeling long sequences.

\section{Factorized Self-Attention}



Sparse Transformers separate the full self-attention operation across several steps of attention, as visualized in Figure \ref{patt:b} and \ref{patt:c}. To motivate our approach, we first perform a qualitative assessment of attention patterns learned by a standard Transformer on an image dataset.


\subsection{Qualitative assessment of learned attention patterns}
We visualized the attention patterns learned by a 128-layer self-attention network on CIFAR-10, and present several examples in Figure \ref{learned_patterns}. Visual inspection showed that most layers had sparse attention patterns across most data points, suggesting that some form of sparsity could be introduced without significantly affecting performance. Several layers (Figure \ref{learned_patterns}c) clearly exhibited global patterns, however, and others exhibited data-dependent sparsity (Figure \ref{learned_patterns}d), both of which would be impacted by introducing a predetermined sparsity pattern into all of the attention matrices.

In this paper, we restricted our investigation to a class of sparse attention patterns that have connectivity between all positions over several steps of attention. These methods can be more efficient than full attention while still providing global context to any given position. We aimed to empirically validate the performance of these factorized patterns on a range of tasks, given that they are unable to learn the exact same mappings as those in Figure \ref{learned_patterns}. We present the formulation of factorized attention below.

\subsection{Factorized self-attention}
A self-attention layer maps a matrix of input embeddings $X$ to an output matrix and is parameterized by a connectivity pattern $S = \{S_1, ..., S_n\}$, where $S_i$ denotes the set of indices of the input vectors to which the $i$th output vector attends. The output vector is a weighted sum of transformations of the input vectors:

\begin{equation}\label{attend_fn}
\mathrm{Attend}(X, S) = \Bigl( \, a(\mathbf{x}_i, S_i) \, \Bigr)_{i \in \{1, ..., n\}}
\end{equation}
\begin{equation}
a(\mathbf{x}_i, S_i) = \mathrm{softmax}\left(\frac{(W_q\mathbf{x}_i)K_{S_i}^T}{\sqrt{d}}\right)V_{S_i}
\end{equation}

\begin{equation}
K_{S_i} = \Bigl( \, W_k \mathbf{x}_j \, \Bigr)_{j \in S_i} \quad V_{S_i} = \Bigl( \, W_v \mathbf{x}_j \, \Bigr)_{j \in S_i}
\end{equation}

Here $W_q$, $W_k$, and $W_v$ represent the weight matrices which transform a given $\mathbf{x}_i$ into a \textit{query}, \textit{key}, or \textit{value}, and $d$ is the inner dimension of the queries and keys. The output at each position is a sum of the values weighted by the scaled dot-product similarity of the keys and queries.

Full self-attention for autoregressive models defines $S_i = \{j : j \le i\}$, allowing every element to attend to all previous positions and its own position. 

Factorized self-attention instead has $p$ separate attention heads, where the $m$th head defines a subset of the indices $A_{i}^{(m)} \subset \{j : j \le i\}$ and lets $S_i = A_{i}^{(m)}$.  We are chiefly interested in \textit{efficient} choices for the subset $A$, where $|A_{i}^{(m)}| \propto \sqrt[p]{n}$.

Additionally, for the time being we consider \textit{valid} choices of $A$, where all input positions are connected to all future output positions across the $p$ steps of attention.

For every $j \le i$ pair, we set every $A$ such that $i$ can attend to $j$ through a path of locations with maximum length $p+1$. Specifically, if $(j, a, b, c, ..., i)$ is the path of indices, then $j \in A_{a}^{(1)}$, $a \in A_{b}^{(2)}$, $b \in A_{c}^{(3)}$, and so forth.

These two criteria allow us keep the ability of Transformers to propagate signals from arbitrary input positions to arbitrary output positions in a constant number of steps, while reducing the total effective computation to $O(n \sqrt[p]{n})$. We also note that softening the validity criterion (for instance, having a series of only locally connected layers) may be a useful inductive bias for certain domains.

In this work, we explore two factorizations for $p=2$, which we describe in the following section, though we note that the same techniques can be easily extended to higher dimensions.

\subsection{Two-dimensional factorized attention}
A natural approach to defining a factorized attention pattern in two dimensions is to have one head attend to the previous $l$ locations, and the other head attend to every $l$th location, where $l$ is the \textit{stride} and chosen to be close to $\sqrt{n}$, a method we call \textit{strided} attention. 

Formally, $A_{i}^{(1)} = \{t, t+1, ..., i \}$ for $t = \mathrm{max}(0, i-l)$ and $A_{i}^{(2)} = \{j : (i - j) \bmod l = 0 \}$. This pattern can be visualized in Figure \ref{patt:b}.

This formulation is convenient if the data naturally has a structure that aligns with the stride, like images or some types of music. For data without a periodic structure, like text, however, we find that the network can fail to properly route information with the strided pattern, as spatial coordinates for an element do not necessarily correlate with the positions where the element may be most relevant in the future.

In those cases, we instead use a \textit{fixed} attention pattern (Figure \ref{patt:c}), where specific cells summarize previous locations and propagate that information to all future cells.

Formally, $A_{i}^{(1)} = \{j : (\lfloor j / l \rfloor = \lfloor i / l \rfloor)\} $, where the brackets denote the floor operation, and $A_{i}^{(2)} = \{j : j \bmod l \in \{t, t + 1, ..., l \}$, where $t = l - c$ and $c$ is a hyperparameter. 

Concretely, if the stride is 128 and $c = 8$, then all future positions greater than 128 can attend to positions 120-128, all positions greater than 256 can attend to 248-256, and so forth.

A fixed-attention pattern with $c = 1$ limits the expressivity of the network significantly, as many representations in the network are only used for one block whereas a small number of locations are used by all blocks. We instead found choosing $c \in \{8, 16, 32\}$ for typical values of $l \in \{128, 256\}$ to perform well, although it should be noted that this increases the computational cost of this method by $c$ in comparison to the strided attention.

Additionally, we found that when using multiple heads, having them attend to distinct subblocks of length $c$ within the block of size $l$ was preferable to having them attend to the same subblock.

In the subsequent section, we describe how to incorporate factorized attention into the Sparse Transformer architecture.

\section{Sparse Transformer}
Here we fully describe the Sparse Transformer architecture, which is a modified version of the Transformer \cite{vaswani2017attention}.

\subsection{Factorized attention heads}
Standard dense attention simply performs a linear transformation of the $\mathrm{attend}$ function defined in Equation \ref{attend_fn}:

\begin{equation}\label{wpmat}
\mathrm{attention}(X) = W_p \cdot \mathrm{attend}(X, S)\end{equation}

where $W_p$ denotes the post-attention weight matrix. The simplest technique for integrating factorized self-attention is to use one attention type per residual block, and interleave them sequentially or at a ratio determined as a hyperparameter:

\begin{equation}\label{sepwpmat}
\mathrm{attention}(X) = W_p \cdot \mathrm{attend}(X, A^{(r \bmod p)})\end{equation}

Here $r$ is the index of the current residual block and $p$ is the number of factorized attention heads.

A second approach is to have a single head attend to the locations of the pixels that both factorized heads would attend to, which we call a \textit{merged} head:

\begin{equation}\label{mergedmat}
\mathrm{attention}(X) = W_p \cdot \mathrm{attend}(X, \bigcup_{m=1}^{p} A^{(m)})\end{equation}

This is slightly more computationally intensive, but only by a constant factor. A third approach is to use multi-head attention \cite{vaswani2017attention}, where $n_{h}$ attention products are computed in parallel, then concatenated along the feature dimension:

\begin{equation}
\mathrm{attention}(X) = W_p \Bigl( \mathrm{attend}(X, A)^{(i)} \Bigr)_{i \in \{1, ..., n_{h}\}}
\end{equation}

Here, the $A$ can be the separate attention patterns, the merged patterns, or interleaved as in Eq. \ref{attend_fn}. Also, the dimensions of the weight matrices inside the $\mathrm{attend}$ function are reduced by a factor of $1/n_{h}$, such that the number of parameters are invariant across values of $n_{h}$.

We typically find multiple heads to work well, though for extremely long sequences where the attention dominates the computation time, it is more worthwhile to perform them one at a time and sequentially.

\subsection{Scaling to hundreds of layers}

\begin{figure}[t]
\centering
\setlength\fboxsep{0pt}
\setlength\fboxrule{0.25pt}
\includegraphics[width=0.40\textwidth]{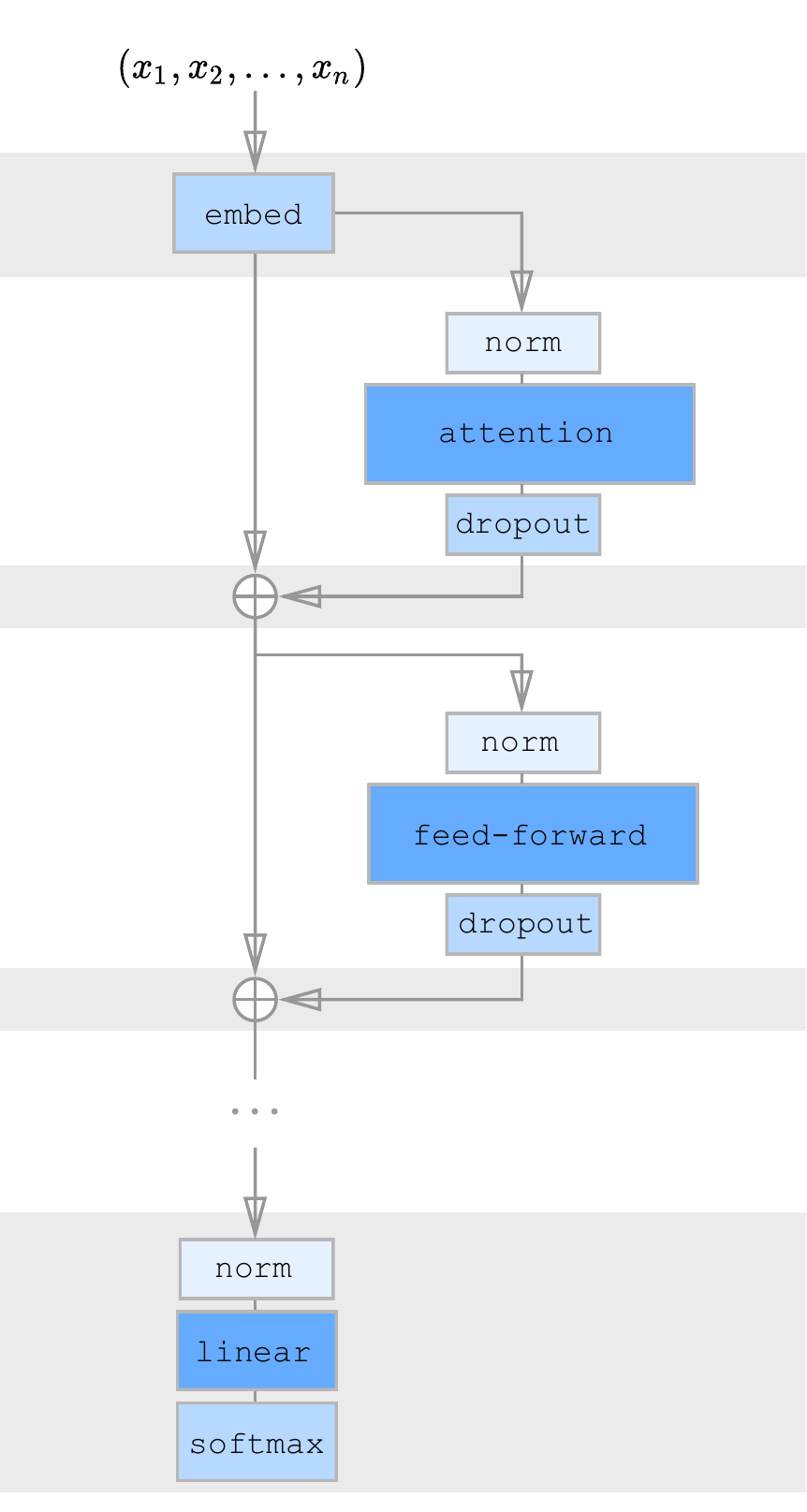}
\caption{Diagram depicting one residual block of the Sparse Transformer. The shaded background indicates tensors which are \textit{checkpointed} \cite{chen2016training} and stored in GPU memory. The other tensors, including the attention weights and feedforward network activations, are recomputed during the calculation of gradients, reducing memory usage substantially. }
\label{gemnetarch}
\end{figure}

We found that Transformers were difficult to train with many layers, as noted by \cite{al2018character}. Instead of incorporating auxillary losses, we adopted the following architectural changes.

First, we use the pre-activation residual block of \cite{He2016}, defining a network of $N$ layers in the following way:
\begin{equation}H_0 = \mathrm{embed}(X, W_e)\end{equation}
\begin{equation}H_k = H_{k-1} + \mathrm{resblock}(H_{k-1})\end{equation}
\begin{equation}y = \mathrm{softmax}(\mathrm{norm}(H_N) W_{out})\end{equation}
where $\mathrm{embed}$ is a function we describe in the next section, $W_{out}$ is a weight matrix, and $\mathrm{resblock}(h)$ normalizes the input to the attention block and a positionwise feedforward network in the following way:
\begin{equation}a(H) = \mathrm{dropout}(\mathrm{attention}(\mathrm{norm}(H)))\end{equation}
\begin{equation}b(H) = \mathrm{dropout}(\mathrm{ff}(\mathrm{norm}(H + a(H))))\end{equation}
\begin{equation}\mathrm{resblock}(H) = a(H) + b(H) \end{equation}

The $\mathrm{norm}$ function denotes Layer Normalization \cite{ba2016layer}, and $\mathrm{ff}(x) = W_2 \, f(W_1x + b_1) + b_2$. Our choice of $f$ is the Gaussian Error Linear Unit \cite{hendrycks2016bridging}, $f(X) = X \odot \, \mathrm{sigmoid}(1.702 \cdot X)$, as used in \cite{radford2018}. The output dimension of $W_1$ is 4.0 times the input dimension, unless otherwise noted.

Observe that $H_N$ is the sum of $N$ applications of functions $a$ and $b$, and thus each function block receives a gradient directly from the output layer . We scale the initialization of $W_2$ and $W_p$ in Eq. \ref{wpmat} by $\frac{1}{\sqrt{2N}}$ to keep the ratio of input embedding scale to residual block scale invariant across values of $N$.

\subsection{Modeling diverse data types}
In addition to the embedding of input symbols, positional embeddings are typically used in Transformers and other location-agnostic architectures to encode the spatial relationships of data \cite{gehring2017convolutional}, \cite{parmar2018image}.

We found using learned embeddings which either encoded the structure of the data or the factorized attention patterns were important for performance of our models. 

We added either $n_{emb} = d_{data}$ or $n_{emb} = d_{attn}$ embeddings to each input location, where $d_{data}$ refers to the number of dimensions of the data, and $d_{attn}$ is the number of dimensions of the factorized attention. If $\mathbf{x}_i$ is the one-hot encoded $i$th element in the sequence, and $\mathbf{o}_{i}^{(j)}$ represents the one-hot encoded position of $\mathbf{x}_i$ in the $j$th dimension ($1 \le j \le n_{emb}$), then:

\begin{equation}
\mathrm{embed}(X, W_e) = \left(\mathbf{x}_iW_e + \sum_{j=1}^{n_{emb}} \mathbf{o}_{i}^{(j)} W_{j} \right)_{\mathbf{x}_i \in X}
\end{equation}

For images, we used data embeddings, where $d_{data} = 3$ for the row, column, and channel location of each input byte. For text and audio, we used two-dimensional attention embeddings, where $d_{attn} = 2$ and the index corresponds to each position's row and column index in a matrix of width equal to the stride.

\begin{figure*}[t]
\centering
\setlength\fboxsep{0pt}
\setlength\fboxrule{0.25pt}
\includegraphics[width=1.0\textwidth]{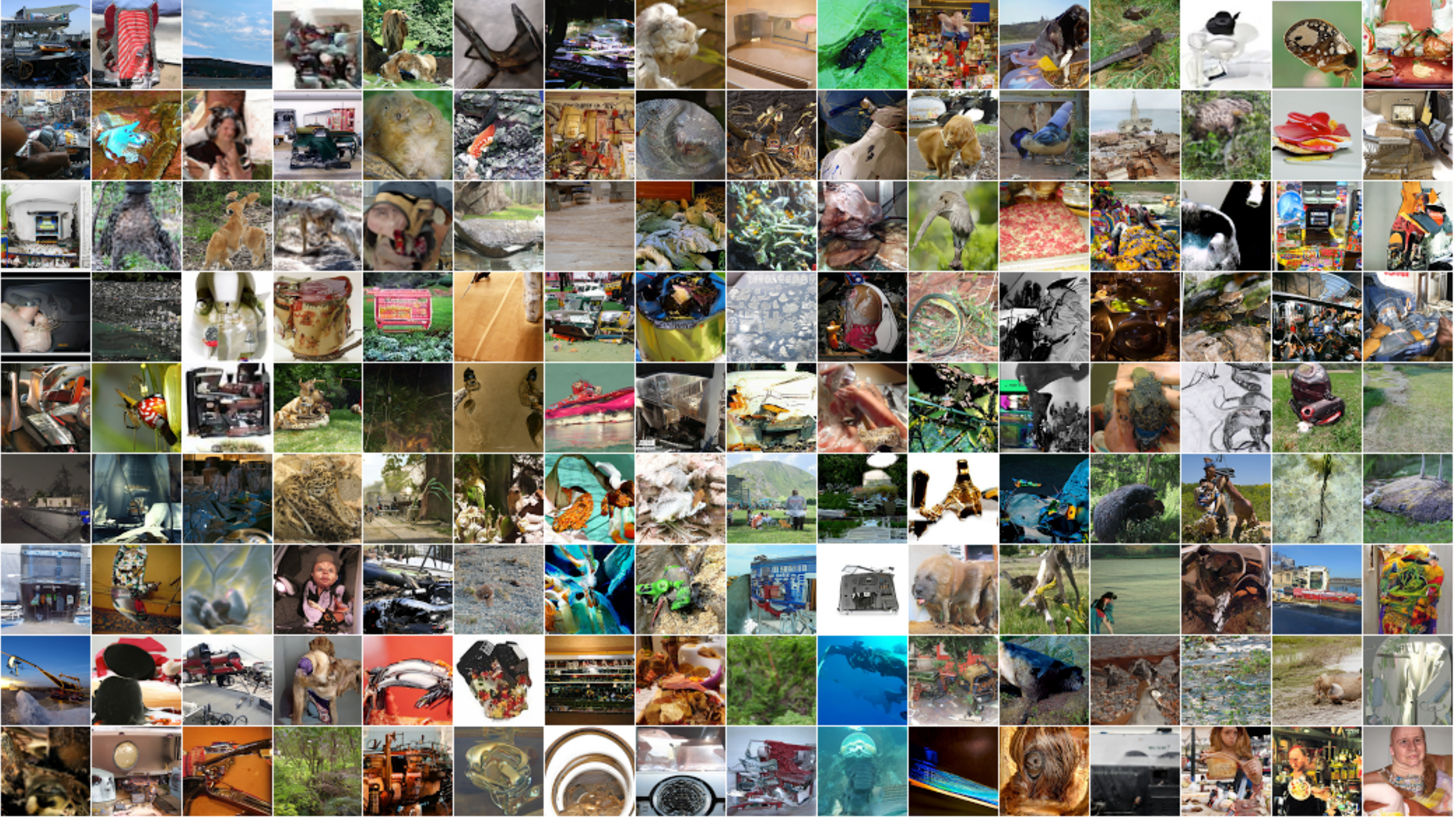}
\caption{Unconditional samples from ImageNet 64x64, generated with an unmodified softmax temperature of 1.0. We are able to learn long-range dependencies directly from pixels without using a multi-scale architecture.}
\label{inet64image}
\end{figure*}

\subsection{Saving memory by recomputing attention weights}
Gradient checkpointing has been shown to be effective in reducing the memory requirements of training deep neural networks \cite{chen2016training}, \cite{gruslys2016memory}. It is worth noting, however, that this technique is particularly effective for self-attention layers when long sequences are processed, as memory usage is high for these layers relative to the cost of computing them.

Using recomputation alone, we are able to train dense attention networks with hundreds of layers on sequence lengths of 16,384, which would be infeasible on modern hardware otherwise.

In our experiments, we recompute the attention and feed-forward blocks during the backwards pass. To simplify our implementation, we do not apply dropout within the attention blocks, as in \cite{vaswani2017attention}, and instead only apply it at the end of each residual addition, as seen in Figure \ref{gemnetarch}.

\subsection{Efficient block-sparse attention kernels}
The sparse attention masks in \ref{patt:b} and \ref{patt:c} can be efficiently computed by slicing out sub-blocks from the query, key, and value matrices and computing the product in blocks. Attention over a local window can be computed as-is, whereas attention with a stride of $k$ can be computed by transposing the matrix and computing a local window. Fixed attention positions can be aggregated and computed in blocks.

In order to ease experimentation, we implemented a set of GPU kernels which efficiently perform these operations. The softmax operation is fused into a single kernel and also uses registers to eliminate loading the input data more than once, allowing it to run at the same speed as a simple nonlinearity. The upper triangle of the attention matrix is never computed, moreover, removing the need for the negative bias term of \cite{vaswani2017attention} and halving the number of operations to be performed.

\subsection{Mixed-precision training}
We store network weights in single-precision floating-point, but otherwise compute network activations and gradients in half-precision, as in \cite{micikevicius2017mixed}. This accelerates our training due to the usage of Tensor Core operations on the V100 GPU. During the gradient calculation, we use dynamic loss scaling to reduce numerical underflow, and we communicate half-precision gradients when averaging across multiple GPUs. When sampling, we cast the queries and keys to single-precision, as the query-key product can sometimes overflow the max value of half-precision.

\section{Training}
We use the Adam optimizer with a linear warmup of 5000 iterations and a gradient clipping of 1.0, both of which we found important for model stability. We use a weight decay penalty of 0.01. We annealed the learning rate according to a cosine decay as in \cite{radford2018}. We train on 8 V100 GPUs unless otherwise noted.

All embeddings are of a constant dimension $d$, usually one of $\{256, 512, 1024\}$. By default, all linear transforms are to the same dimension, with the exception of the feed-forward network, which projects the input to $4d$, unless we use ``half-size'' transformations, where it is $2d$. Additionally, sometimes we halve the size of the query and key transformations.

We initialize the token embedding $W_e$ from $\mathcal{N}(0, \frac{0.125}{\sqrt{d}})$ and the position embeddings from $\mathcal{N}(0, \frac{0.125}{\sqrt{dn_{emb}}})$. Within the attention and feedforward components, all biases are initialized to 0 and all weights are initialized from $\mathcal{N}(0, \frac{0.125}{\sqrt{d_{in}}})$ where $d_{in}$ is the fan-in dimension. The weight matrix for the output logits was initialized to 0.

\section{Experiments}
We empirically test our architecture on density modeling tasks including natural images, text, and raw audio. A summary of the results is available in Table \ref{all_results}. We found that, in addition to running significantly faster than full attention, sparse patterns also converged to lower error, as shown in Table \ref{compare_attn}. This may point to a useful inductive bias from the sparsity patterns we introduced, or an underlying optimization issue with full attention.

\begin{table}[t]
\caption{Summary of our findings for density modeling tasks. Results are reported in bits per byte, which is equivalent to bits per dim for image tasks. M refers to millions of parameters.}
\label{all_results}
\vskip 0.15in
\begin{center}
\begin{small}
\begin{tabular}{lcccr}
\toprule
Model & Bits per byte\\
\quad \\

\textbf{CIFAR-10}\\
\midrule
PixelCNN \cite{oord2016pixel} & 3.03\\
PixelCNN++ \cite{salimans2017pixelcnn++} & 2.92\\
Image Transformer \cite{parmar2018image} & 2.90\\
PixelSNAIL \cite{chen2017pixelsnail} & 2.85\\
\textbf{Sparse Transformer 59M (strided)} & \textbf{2.80} \\

\quad \\
\textbf{Enwik8}\\
\midrule
Deeper Self-Attention \cite{al2018character} & 1.06\\
Transformer-XL 88M \cite{dai2018transformer} & 1.03 \\
Transformer-XL 277M \cite{dai2018transformer} & \textbf{0.99}\\
\textbf{Sparse Transformer 95M (fixed)} & \textbf{0.99}\\

\quad \\
\textbf{ImageNet 64x64}\\
\midrule
PixelCNN \cite{oord2016pixel} & 3.57 \\
Parallel Multiscale \cite{reed2017parallel} & 3.7 \\
Glow \cite{kingma2018glow} & 3.81 \\
SPN 150M \cite{menick2018generating} & 3.52 \\
\textbf{Sparse Transformer 152M (strided)} & \textbf{3.44} \\

\quad \\
\textbf{Classical music, 5 seconds at 12 kHz} \\
\midrule
Sparse Transformer 152M (strided) & \textbf{1.97} \\

\bottomrule
\end{tabular}
\end{small}
\end{center}
\vskip -0.1in
\end{table}

\begin{table}[h]
\caption{Sparse patterns showed increased speed and also better loss on the datasets where we could compare both, which may point to a useful inductive bias in the patterns we learned or an underlying optimization issue with full attention.}
\label{compare_attn}
\vskip 0.15in
\begin{center}
\begin{small}
\begin{tabular}{lcccr}
\toprule
Model & Bits per byte & Time/Iter\\
\quad \\
\textbf{Enwik8 (12,288 context)} \\
\midrule
Dense Attention & 1.00 & 1.31\\
Sparse Transformer (Fixed) & \textbf{0.99} & 0.55 \\
Sparse Transformer (Strided) & 1.13 & 0.35 \\
\quad \\
\textbf{CIFAR-10 (3,072 context)} \\
\midrule
Dense Attention & 2.82 & 0.54 \\
Sparse Transformer (Fixed) & 2.85 & 0.47 \\
Sparse Transformer (Strided) & \textbf{2.80} & 0.38 \\

\bottomrule
\end{tabular}
\end{small}
\end{center}
\vskip -0.2in
\end{table}

\begin{table}[h]
\caption{We observe increased compression of Enwik8 with longer contexts, suggesting the Sparse Transformer can effectively incorporate long-term dependencies.}
\label{compare_ctx}
\begin{center}
\begin{small}
\begin{tabular}{lcccr}
\toprule
Minimum context length during evaluation & Bits per byte\\
\midrule
6,144 tokens & 0.9952\\
9,216 tokens & 0.9936 \\
10,752 tokens & 0.9932 \\
11,904 tokens & 0.9930 \\
12,096 tokens & 0.9922 \\
12,160 tokens & \textbf{0.9908} \\

\bottomrule
\end{tabular}
\end{small}
\end{center}
\vskip -0.12in
\end{table}

\subsection{CIFAR-10}

We train strided Sparse Transformers on CIFAR-10 images represented as sequences of 3072 bytes. Models have 2 heads, 128 layers, $d$ = 256, half-size feedforward network and query-key projections, and are trained for 120 epochs with a learning rate of 0.00035 and a dropout rate of 0.25 until validation error stops decreasing.

We use 48000 examples for training and 2000 examples for validation, evaluating the performance of our best models on the test set. The model achieves \textbf{2.80} bits per dim ($2.798 \pm 0.004$ over seeds 1, 2, 3) versus the previous $2.85$ state of the art \cite{chen2017pixelsnail}. We also compare performance of different attention patterns in Table \ref{compare_attn}. The strided attention reaches the lowest error in the shortest amount of time, surpassing the error of dense attention at 2.82 bits per dim.

\subsection{Text}
In order to assess Sparse Transformers on datasets without a strong two-dimensional structure, we trained models on the EnWik8 dataset, which represents the first $10^{8}$ bytes of Wikipedia and contains a great degree of variability in periodic structure. We trained with a context length of 12,288, which is longer than previous approaches.

We trained on the first 90 million tokens and reserved the last 10 million for validation and test. We used 30-layer fixed Sparse Transformers with 8 heads, $d$ = 512, and a dropout rate of $0.40$. We trained for 80 epochs until validation loss stopped decreasing. We used a stride of 128, $c = 32$, and merged the factorized attention heads.

Our best model reached \textbf{0.99} bits per dim ($0.992 \pm 0.001$ over seeds 1, 2, 3), surpassing the 1.03 state-of-the-art for a similarly-sized Transformer-XL  \cite{dai2018transformer} and matching the 0.99 of a model trained with more than double the number of parameters. Strided attention failed to do well on this dataset, whereas fixed patterns were able to recover and surpass the performance of dense attention, as listed in Table \ref{compare_attn}.

Additionally, during evaluation of the test set, we modified the minimum context length the network could use by evaluating fewer tokens in parallel. We saw monotonic increases in performance with more tokens used, up to 12,160 out of the 12,288 tokens used for training (see Table \ref{compare_ctx}), which suggests the network is effectively incorporating long-term dependencies.

\subsection{ImageNet 64x64}

In order to test the ability of the model to learn long range dependencies and scale to a large dataset, we train on the version of downsampled ImageNet released by \cite{oord2016pixel} and evaluate on the validation set. We used a 48 layer strided Sparse Transformer with 16 attention heads and $d$ = 512, totaling 152 million parameters. We used a stride of 128, a dropout of 0.01, and trained for 70 epochs, which took 7 days on 64 V100 GPUs.

Our model achieves a loss of \textbf{3.44} bits per dim (3.437 across 1 run), in comparison to the previous 3.52 \cite{menick2018generating}. 

Additionally, we generate unconditional samples (Figure \ref{inet64image}) at an unmodified softmax temperature of 1.0, from the model and from one trained with twice the layers (300M parameters total). We include here samples from the 300M parameter model. On visual assessment we find no artifacts from the sparsity patterns and see evidence of long-term structure in most images. 

\subsection{Classical music from raw audio}

To test the extent to which Sparse Transformers are able to scale to very long contexts, we trained models on the classical music dataset released by \cite{dieleman2018challenge}. As details of the dataset processing are unavailable, we omit any direct comparison to other work and instead study what size of Sparse Transformer we can train with increasing context size. For each sequence length, we attempted to train the largest model which could entirely fit into 16GB V100 accelerators without model parallelism.

Overall, we found that increasing the sequence length by a factor of 4 requires a reduction in model capacity of approximately $4 \sqrt{4} = 8$. Thus we found we could use factorized self-attention on sequences over 1 million timesteps long, albeit with extremely few parameters (3 million).

Samples are available for sequences of length 65,536, which correspond to around 5 seconds of generated audio at 12kHz. The samples clearly demonstrate global coherence over the sampled period, and exhibit a variety of play styles and tones, swapping from rhythmic playing to forceful. To listen to samples, visit \url{https://openai.com/blog/sparse-transformer}. Sample quality quickly degrades for greater sequence lengths due to reduced model capacity. 

\begin{table}[h]
\caption{Performance of a strided Sparse Transformer on a classical audio dataset ($\mu$-law encoded at 12 kHz) as a function of sequence length and model size.}
\label{audio}
\vskip 0.15in
\begin{center}
\begin{small}
\begin{tabular}{lcccr}
\toprule
Sequence length & Parameters & Bits per byte \\
\midrule
65,536 & 152M & 1.97 \\
262,144 & 25M & 2.17 \\
1,048,576 & 3M & 2.99 \\

\bottomrule
\end{tabular}
\end{small}
\end{center}
\vskip -0.1in
\end{table}

\section{Conclusion}
We introduced Sparse Transformers and showed they attain equivalent or better performance on density modeling of long sequences than standard Transformers while requiring significantly fewer operations. This performance is state-of-the-art in images and text and is easily adaptable to raw audio. The model demonstrates usage of long-term context and generates globally coherent samples.

\section{Acknowledgements}
We would like to thank Ashish Vaswani for insightful discussions during the genesis of the project. We also thank Joshua Meier and Mark Chen for helpful discussions, and Johannes Otterbach, Prafulla Dhariwal, and David Luan for feedback on drafts of this paper.

\nocite{chen2017pixelsnail}
\nocite{salimans2017pixelcnn++}

\bibliography{sparse_transformers}
\bibliographystyle{icml2019}

\end{document}